\documentclass[conference]{IEEEtran}
\IEEEoverridecommandlockouts
\usepackage{cite}
\usepackage{amsmath,amssymb,amsfonts}
\usepackage{graphicx}
\usepackage{booktabs}
\usepackage{multirow}
\usepackage{makecell}
\usepackage{textcomp}
\usepackage{url}
\def\BibTeX{{\rm B\kern-.05em{\sc i\kern-.025em b}\kern-.08em T\kern-.1667em\lower.7ex\hbox{E}\kern-.125emX}}

\begin{document}

\title{MAOL: Morphology-Aware Ordinal Learning for Fine-Grained Industrial Defect Severity Grading}

\author{
\resizebox{0.86\textwidth}{!}{%
\begin{tabular}{ccc}
Zhaoyang Wang & Haiyong Chen & Binyi Su\\
\textit{Hebei University of Technology} & \textit{Hebei University of Technology} & \textit{Hebei University of Technology}\\
Tianjin, China & Tianjin, China & Tianjin, China\\
895056337@qq.com & haiyong.chen@hebut.edu.cn & subinyi@hebut.edu.cn
\end{tabular}}\\[0.35em]
\resizebox{0.98\textwidth}{!}{%
\begin{tabular}{cccc}
Kun Liu & Kun Wang & Xianen Zhou & ATIK SHAHARIAR\\
\textit{Hebei University of Technology} & \textit{Yingli Energy Development Co. Ltd} & \textit{Hunan University} & \textit{Hebei University of Technology}\\
Tianjin, China & Baoding, China & Changsha, China & Tianjin, China\\
2009069@hebut.edu.cn & 13931278136@163.com & zhouxianen@hnu.edu.cn & atik.shahariar.5@gmail.com
\end{tabular}}
\thanks{This work was supported in part by the National Natural Science Foundation of China under Grant 62473127, the National Key Research and Development Program of China under Grant 2024YFB3310904, the S\&T Program of Hebei Province 242Q4302Z, the Beijing-Tianjin-Hebei Basic Research Cooperation Special Project F2024202102, and the China Scholarship Council 202506700027.}
}

\maketitle

\begin{abstract}
Fine-grained defect severity grading is essential for industrial inspection, yet remains challenging due to the ordinal nature of severity labels, the strong dependence on morphology-related cues, and the train--test discrepancy between clean annotated instances and noisy predicted instances in two-stage pipelines. We propose MAOL, a Morphology-Aware Ordinal Learning framework for fine-grained industrial defect severity grading. MAOL formulates severity grading as an instance-level ordinal learning task, incorporates explicit morphological features to enhance representation learning, introduces class-conditional adaptive ordinal thresholds to model defect-specific grading boundaries, and employs prediction-aware training via localization perturbation to improve robustness to imperfect predicted instances. Extensive experiments under both clean-ROI and predicted-instance settings demonstrate that MAOL consistently outperforms rule-based methods, nominal classification models, and existing ordinal baselines, especially in the predicted-instance setting. The proposed approach ranked third in the IDA 2026 Challenge on Fine-Grained Severity Grading for High-Precision Manufacturing.
\end{abstract}


\begin{figure}[!t]
\centering
\includegraphics[width=3.4in]{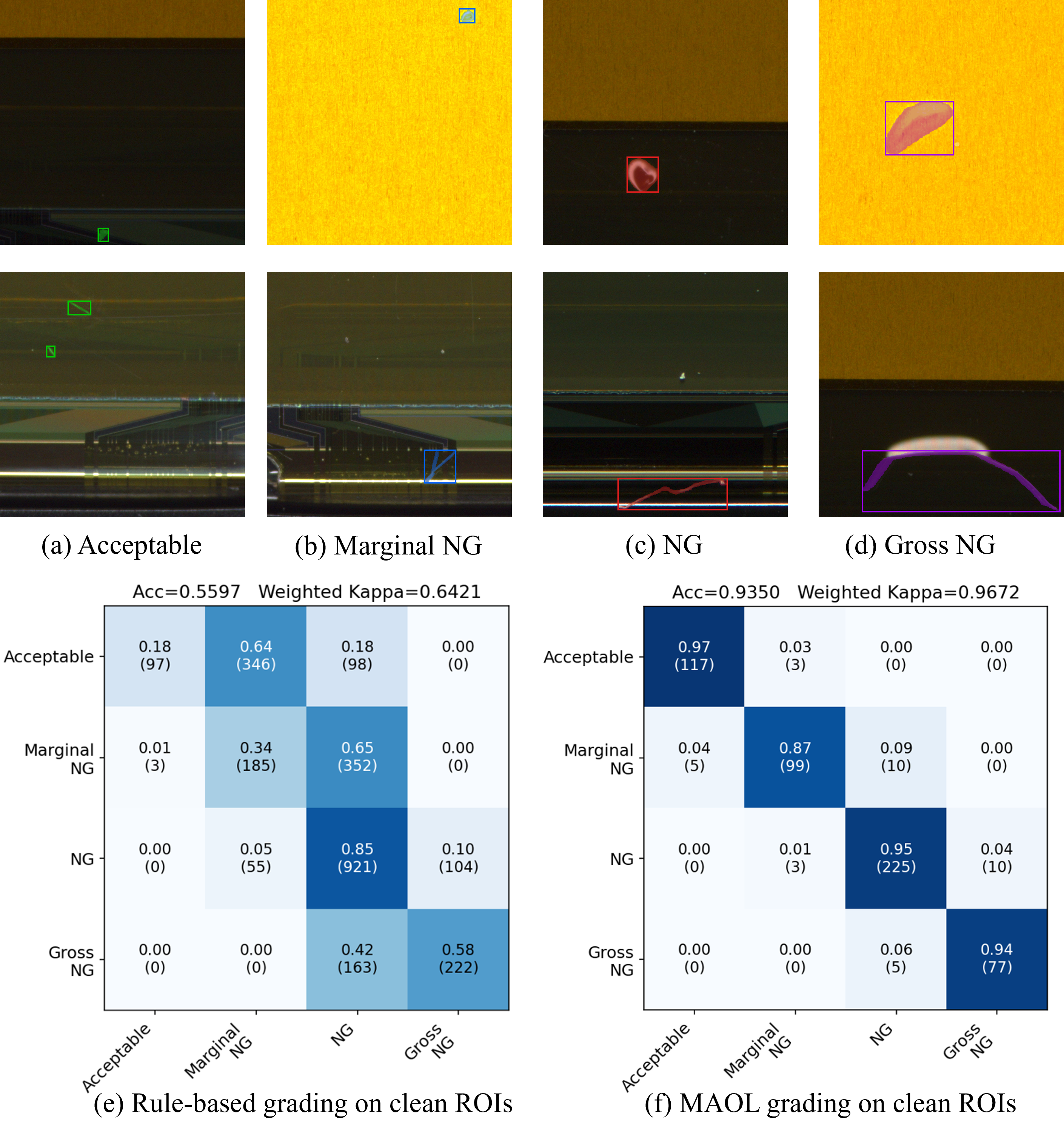}
\caption{Illustration of fine-grained industrial defect severity grading. (a)--(d) Representative examples across severity levels. (e) Confusion matrix of a rule-based method under the clean-ROI setting. (f) Confusion matrix of the proposed MAOL under the same setting.}
\label{fig_1}
\end{figure}

\section{Introduction}
In high-precision manufacturing, industrial visual inspection requires not only detecting defects but also assessing their severity levels to support quality screening, rework decision-making, and production control \cite{11209012,cpdd}. Compared with conventional binary defect judgment, severity grading provides more fine-grained quality information and plays an increasingly important role in intelligent inspection systems. In practical scenarios, defects of different severity levels correspond to different handling strategies, making accurate and reliable severity prediction a critical yet challenging problem \cite{li2025defect}.

A straightforward solution is to design rule-based grading systems using morphology-related statistics such as defect area, shape, or contrast. However, such methods heavily rely on handcrafted thresholds and expert knowledge, and often generalize poorly to complex industrial scenarios. As illustrated in Fig.~\ref{fig_1}(e), even under the clean-ROI setting, a rule-based method exhibits a scattered confusion pattern and achieves only 0.5597 Accuracy and 0.6421 Quadratic Weighted Kappa (QWK) \cite{bendavid2008ordinal}, indicating its limited capability in fine-grained severity discrimination.

A more common alternative is to formulate severity grading as a classification problem on cropped Regions of Interest (ROIs). However, this formulation is also suboptimal. Severity labels are inherently ordered, and treating them as independent categories ignores the relative relationships between different severity levels. As a result, classification-based models fail to distinguish minor adjacent-level errors from large cross-level mistakes.

Another critical challenge arises from the train--test discrepancy in two-stage inspection pipelines. In practice, severity grading is performed on detector-predicted ROIs rather than perfectly aligned ground-truth regions. These predicted instances inevitably introduce localization errors, background contamination, and incomplete defect regions, which significantly degrade downstream grading performance. Therefore, an effective severity grading framework should jointly model ordinal label structure, preserve morphology-related information, account for category-specific grading differences, and remain robust to imperfect predicted-instance inputs.

To address these challenges, we propose MAOL, a Morphology-Aware Ordinal Learning framework for fine-grained industrial defect severity grading. MAOL formulates severity grading as an instance-level ordinal learning problem and is designed as a problem-aligned solution to the above challenges. Specifically, it incorporates explicit morphology descriptors to preserve structural cues, introduces class-conditional adaptive ordinal thresholds to model category-dependent grading boundaries, and adopts prediction-aware training via localization perturbation to improve robustness to noisy predicted ROIs.

Extensive experiments under both clean-ROI and predicted-instance settings demonstrate the effectiveness of MAOL. As shown in Fig.~\ref{fig_1}(f), MAOL achieves 0.935 Acc. and 0.9672 QWK under the clean-ROI setting, with a substantially cleaner confusion pattern than the rule-based baseline. Quantitative results further show that MAOL consistently outperforms rule-based methods, classification-based models, and existing ordinal baselines, with particularly significant improvements under predicted-instance conditions. The main contributions of this work are summarized as follows:
\begin{itemize}
\item We formulate fine-grained industrial defect severity grading as an instance-level ordinal learning problem and propose MAOL, a unified framework that explicitly models label ordering.
\item We introduce a morphology-aware ordinal learning scheme with class-conditional adaptive thresholds to capture structural cues and category-specific grading criteria.
\item We propose a prediction-aware training strategy to bridge the gap between clean annotations and noisy predicted instances, improving robustness in practical two-stage inspection pipelines.
\item Extensive experiments in GT and predicted ROI settings demonstrate the superiority of the proposed method. Our MAOL achieves a final score of 0.7965 (S2 metric) in the IDA 2026 Challenge, ranking third among all participants.
\end{itemize}

\section{Related Work}
\subsection{Industrial Defect Inspection and Severity Assessment}
Industrial defect detection and segmentation have long been core research topics in industrial visual inspection. With the rapid development of deep learning, related methods have evolved from traditional hand-crafted features and shallow classifiers to convolutional neural network and Transformer-based frameworks for defect detection, segmentation, and anomaly localization \cite{11397733,10292540}. In contrast, defect severity assessment and quality-oriented evaluation have received comparatively less attention, although they are more challenging downstream tasks that require models not only to identify defects, but also to distinguish fine-grained differences among multiple severity levels.

Existing studies fall mainly into two categories. The first category relies on manually designed rules, where defects are quantified according to statistics such as area, shape, contrast, grayscale intensity, or effective defect region, and then mapped to severity levels using predefined formulas or thresholds \cite{liu2025quantitative}. Although such methods are relatively interpretable, they usually depend heavily on human experience and generalize poorly to complex industrial scenarios with large appearance variations and cross-category differences. The second category attempts to directly predict defect severity or product quality using deep neural networks, sometimes jointly with detection or segmentation. Representative examples include the multi-branch U-Net \cite{unet} for joint defect type and severity segmentation proposed by Neven and Goedem\'e \cite{neven2021multi}, the DeepLabv3+ based framework \cite{deeplabv3plus} for sewer defect detection and severity quantification introduced by Zhou et al. \cite{zhou2022automatic}, the defect-attribute-based quality assessment network proposed by Zhang et al. \cite{zhang2025industrial}, and the two-stage defect detection and severity grading model developed by Tian et al. \cite{tian2025ddspnet}.

Although these methods demonstrate the importance of severity-aware inspection, most of them still formulate severity grading as a standard classification or application-specific scoring problem, and only limited work explicitly considers the ordered relationship among severity labels \cite{tian2025ddspnet}. In contrast, this paper focuses on instance-level, fine-grained industrial defect severity grading with ordinal label structure, while further considering the impact of predicted-instance noise in real two-stage systems.

\subsection{Ordinal Regression and Ordered Visual Recognition}
\begin{figure}[!t]
\centering
\includegraphics[width=3.4in]{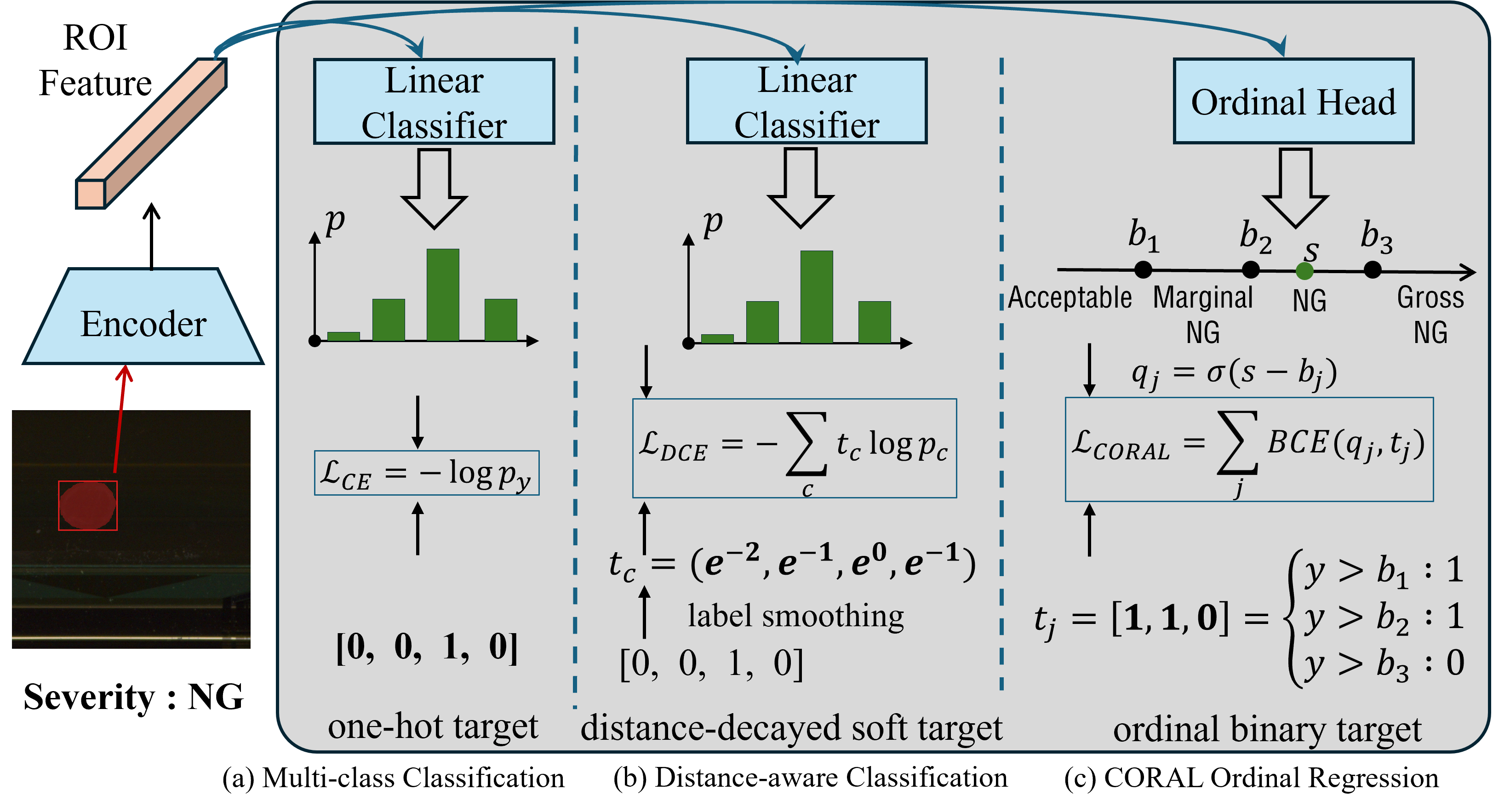}
\caption{Comparison of three formulations for severity prediction using ROI features: (a) multi-class classification with one-hot supervision, (b) distance-aware classification with class-distance-dependent penalties, and (c) CORAL ordinal regression with cumulative threshold modeling and ordinal binary targets.}
\label{fig_2}
\end{figure}

Ordinal regression explicitly models label ordering and is therefore more suitable than standard classification for severity grading \cite{wang2025survey}. It distinguishes adjacent-level errors from large cross-level errors, which is desirable when mistake severities are not equally penalized. Related ideas have also been explored in distance-aware classification, where penalties increase with class distance \cite{polat2025class}. Fig.~\ref{fig_2} illustrates the differences among these formulations.

In deep ordinal learning, representative methods such as CORAL (COnsistent RAnk Logits) and CORN (Conditional Ordinal Regression for Neural networks) have shown strong effectiveness by explicitly modeling rank structure \cite{coral,corn}. In particular, CORAL enforces rank consistency across ordinal thresholds, ensuring that predictions follow a valid monotonic ordering. Similar grading-oriented ideas have also been explored in applications such as defect and agricultural quality assessment \cite{al2024external,knott2025weakly}. However, explicit ordinal learning remains underexplored in industrial defect severity grading, where most existing methods rely on handcrafted rules or distance-aware classification. In this work, we build upon CORAL and further extend it with morphology-aware representation, class-conditional grading boundaries, and prediction-aware training.

\section{Method}
We address fine-grained industrial defect severity grading in a two-stage inspection pipeline and propose MAOL (Morphology-Aware Ordinal Learning), a unified framework for practical deployment. MAOL formulates severity grading as an instance-level ordinal prediction task and is designed to handle morphology-sensitive cues and deployment-time discrepancies. The overall architecture is illustrated in Fig.~\ref{fig_3}.

\subsection{Overall Framework}
Given an input image $I$, an upstream instance segmentation model first detects and segments defect instances. For the $i$-th instance, let $r_i$ denote its region representation and $c_i$ its defect category. An ROI is extracted from $r_i$, resized to $64\times64$ pixels, and fed into a downstream severity grading network, which predicts the severity label $\hat{y}_i \in \{0,1,\ldots,K-1\}$, where $K$ is the number of ordered severity levels. In our setting, $K=4$, corresponding to \emph{Acceptable}, \emph{Marginal NG}, \emph{NG}, and \emph{Gross NG}.

\begin{figure*}[!t]
\centering
\includegraphics[width=0.90\textwidth]{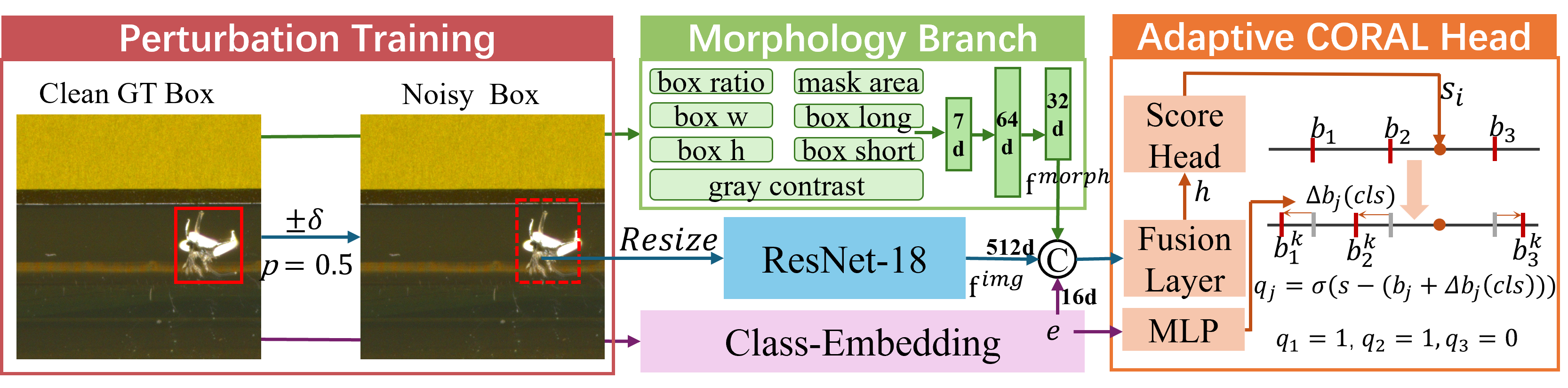}
\caption{Overview of the proposed MAOL framework, illustrating perturbation-aware training, morphology-aware representation, and adaptive ordinal prediction.}
\label{fig_3}
\end{figure*}

We formulate severity grading as an instance-level ordinal prediction problem. As illustrated in Fig.~\ref{fig_3}, the proposed MAOL framework contains three key components: perturbation-aware training on the ROI image branch, morphology- and category-aware representation learning, and an adaptive CORAL head for ordinal severity grading. Specifically, the resized ROI is processed by a ResNet18 \cite{Resnet} image encoder, while explicit morphology descriptors are extracted in parallel from the defect instance. These features are then fused with category information and passed to the adaptive ordinal head for final severity grading.

\subsection{Adaptive CORAL Head}
Severity labels are inherently ordered. Therefore, we adopt CORAL as the ordinal prediction head and further make its decision boundaries category-adaptive. For each instance with severity label $y_i \in \{0,1,\ldots,K-1\}$, we convert it into an ordinal binary target vector of length $K-1$:
\begin{equation}
\mathbf{t}_i=[\mathbb{I}(y_i>0),\mathbb{I}(y_i>1),\ldots,\mathbb{I}(y_i>K-2)].
\end{equation}
For the four severity levels in our task, the encoding becomes
\begin{equation}
0\rightarrow[0,0,0],\quad 1\rightarrow[1,0,0],\quad 2\rightarrow[1,1,0],\quad 3\rightarrow[1,1,1].
\end{equation}

Let $\mathbf{h}_i\in\mathbb{R}^{d}$ denote the fused representation of the $i$-th instance, and let $\mathbf{e}_i$ denote the learnable embedding of its defect category. The adaptive CORAL head first computes a scalar severity score
\begin{equation}
s_i=\mathbf{w}_s^{\top}\mathbf{h}_i,
\end{equation}
which represents the position of the instance on a shared latent severity axis. A globally shared bias vector $\mathbf{b}^{g}\in\mathbb{R}^{K-1}$ defines the base ordinal thresholds, while a class-conditional offset is generated from the category embedding:
\begin{equation}
\Delta\mathbf{b}(c_i)=\mathrm{MLP}(\mathbf{e}_i)\in\mathbb{R}^{K-1}.
\end{equation}
The final rank logits are then defined as
\begin{equation}
z_{i,j}=s_i+b^{g}_{j}+\Delta b_j(c_i),\qquad j=0,\ldots,K-2.
\end{equation}
We optimize the ordinal prediction using binary cross-entropy over all rank logits:
\begin{equation}
\mathcal{L}_{\mathrm{coral}}=\frac{1}{N(K-1)}\sum_{i=1}^{N}\sum_{j=0}^{K-2}\mathrm{BCE}\!\left(\sigma(z_{i,j}),t_{i,j}\right),
\end{equation}
where $N$ is the batch size and $\sigma(\cdot)$ is the sigmoid function. During inference, the final severity label is obtained by counting the number of passed thresholds:
\begin{equation}
\hat{y}_i=\sum_{j=0}^{K-2}\mathbb{I}\!\left(\sigma(z_{i,j})>0.5\right).
\end{equation}
This design jointly models a shared ordinal severity axis and category-adaptive decision boundaries, thereby enabling more accurate capture of category-specific grading criteria than fixed-threshold ordinal predictors.

\subsection{Morphology Branch and Class Embedding}
ROI appearance alone may not fully capture the structural cues that are highly correlated with defect severity. In industrial scenarios, severity is strongly related to geometric and intensity-related properties such as defect size, shape, and contrast. Moreover, such cues can be distorted by ROI resizing, making them difficult to learn purely from appearance features.

To address this issue, we introduce a morphology branch together with a learnable class embedding module. As illustrated in Fig.~\ref{fig_3}, the cropped ROI is resized to $64\times64$ and encoded by a ResNet18 backbone to obtain an image feature $\mathbf{f}^{\mathrm{img}}_i\in\mathbb{R}^{512}$. In parallel, a set of morphology descriptors capturing defect size, geometry, and contrast is extracted and projected into a low-dimensional embedding $\mathbf{f}^{\mathrm{morph}}_i$ using a lightweight encoder.

In addition, the defect category $c_i$ is mapped to a learnable embedding $\mathbf{e}_i=\mathrm{Emb}(c_i)$. The final representation is obtained by feature fusion:
\begin{equation}
\mathbf{h}_i=\phi\!\left([\mathbf{f}^{\mathrm{img}}_i;\mathbf{f}^{\mathrm{morph}}_i;\mathbf{e}_i]\right),
\end{equation}
where $\phi(\cdot)$ denotes a learnable fusion module. This design enables the model to jointly leverage appearance features, scale-consistent morphology cues, and category-level context, leading to more reliable ordinal severity grading.

\subsection{Perturbation-aware Training}
In practical two-stage inspection systems, severity grading operates on detector-predicted ROIs rather than perfectly aligned ground-truth regions, leading to a distribution shift caused by localization errors such as offsets, scale variations, truncations, and background contamination.

To model this deployment-time discrepancy, we introduce a perturbation-aware training strategy on the ROI image branch. As illustrated in Fig.~\ref{fig_3}, during training, the GT box is randomly perturbed with probability $p=0.5$ by applying small independent offsets to its boundaries, resulting in a perturbed box $\tilde{b}_i$. The perturbation is constrained by an IoU range $[\tau_{\min},\tau_{\max}]$ (0.75--0.90) to ensure realistic deviations. The perturbed box is used to crop the ROI for the image branch, while morphology features are computed from the original instance. Optionally, contrast-related morphology cues can be recomputed from the perturbed ROI. This design simulates detector-induced noise during training and improves robustness to localization errors in real-world deployment, without altering the severity labels.

\section{Experiments}
\subsection{Datasets And Experimental Details}
We evaluate MAOL on the severity grading track of an industrial defect inspection benchmark. Each defect instance is annotated with a category, a segmentation mask, and an ordinal severity label from four ordered levels: \emph{Acceptable}, \emph{Marginal NG}, \emph{NG}, and \emph{Gross NG}. The dataset is split at the image level into training and validation sets, as summarized in Table~\ref{tab:split}. The training split contains 924 images and 1992 instances, while the validation split contains 230 images and 554 instances. The severity distribution is imbalanced, making the task challenging.

\begin{table}[t]
\centering
\caption{Training and validation split statistics for severity grading.}
\label{tab:split}
\resizebox{\columnwidth}{!}{%
\begin{tabular}{lrrrrrr}
\toprule
Split & \#Img. & \#Inst. & Acceptable & Marginal NG & NG & Gross NG\\
\midrule
Train & 924 & 1992 & 421 & 426 & 842 & 303\\
Val & 230 & 554 & 120 & 114 & 238 & 82\\
\bottomrule
\end{tabular}}
\end{table}

We consider two settings: Clean ROI, where ROIs are cropped from GT annotations, and Predicted ROI, where ROIs are obtained from a first-stage detector. The latter is more realistic but more difficult due to localization errors. In this setting, all morphology descriptors are likewise computed from the detector outputs.

The first-stage detector is YOLOv8X-seg \cite{yolov8}, trained for 300 epochs with input size $512\times512$ and batch size 32. For severity grading, each ROI is resized to $64\times64$, and ResNet18 is used as the image backbone. MAOL is trained for 300 epochs with batch size 64. All experiments are conducted on an NVIDIA RTX 4090 GPU. Unless otherwise specified, the best model is selected by validation QWK. Performance is reported using Accuracy (Acc.), Macro-average Accuracy (Macro Acc.), and Quadratic Weighted Kappa (QWK) \cite{bendavid2008ordinal}.

\subsection{Results And Comparisons}
\subsubsection{Quantitative Results}

\begin{table*}[t]
\centering
\caption{Overall comparison of different severity grading methods under clean and predicted ROI settings.}
\label{tab:overall_comparison}
\begin{tabular}{lcccccc}
\toprule
\multirow{2}{*}{Method} & \multicolumn{3}{c}{Clean ROI (GT)} & \multicolumn{3}{c}{Predicted ROI}\\
\cmidrule(lr){2-4}\cmidrule(lr){5-7}
& Acc. & Macro Acc. & QWK & Acc. & Macro Acc. & QWK\\
\midrule
Rule-based & 56.0 & 49.8 & 64.2 & 29.7 & 31.8 & 27.4\\
Classification Head & 70.7 & 67.3 & 77.7 & 34.0 & 45.4 & 34.5\\
Distance-aware Head & 78.7 & 76.8 & 87.3 & 35.9 & 46.9 & 46.6\\
CORAL \cite{coral} & 80.8 & 81.2 & 88.8 & 55.8 & 42.1 & 51.1\\
CORN \cite{corn} & 69.8 & 68.5 & 78.7 & 39.3 & 31.7 & 25.9\\
MAOL (Ours) & 93.5 & 94.4 & 96.7 & 84.8 & 81.2 & 88.9\\
\bottomrule
\end{tabular}
\end{table*}

The first-stage detector achieves 86.5\% mAP@0.5 (box) and 84.1\% mAP@0.5 (mask) on the val set. Table~\ref{tab:overall_comparison} summarizes the comparison of severity grading methods under the Clean ROI and Predicted ROI settings. First, learning-based methods consistently outperform the rule-based baseline, with a much larger gap under Predicted ROI (29.7\% Acc. / 27.4 QWK), indicating its sensitivity to localization errors. Second, all methods degrade under Predicted ROI due to detector-induced noise such as offsets, scale mismatch, and background contamination. Third, distance-aware classification improves over standard classification, confirming the benefit of incorporating class-distance information. CORAL further achieves the best performance among the baselines, demonstrating the advantage of explicitly modeling ordinal structure. In contrast, CORN performs worse, especially under Predicted ROI (25.9 QWK), likely due to its sensitivity to limited data, class imbalance, and ROI noise. Finally, MAOL achieves the best results by a large margin in both settings. Notably, under Predicted ROI, it surpasses the baseline by +51.1 Acc., +46.0 Macro Acc., and +61.5 QWK, highlighting the effectiveness of morphology-aware representation, adaptive thresholds, and prediction-aware training.

\subsubsection{Qualitative Analysis}
\begin{figure}[!t]
\centering
\includegraphics[width=3.4in]{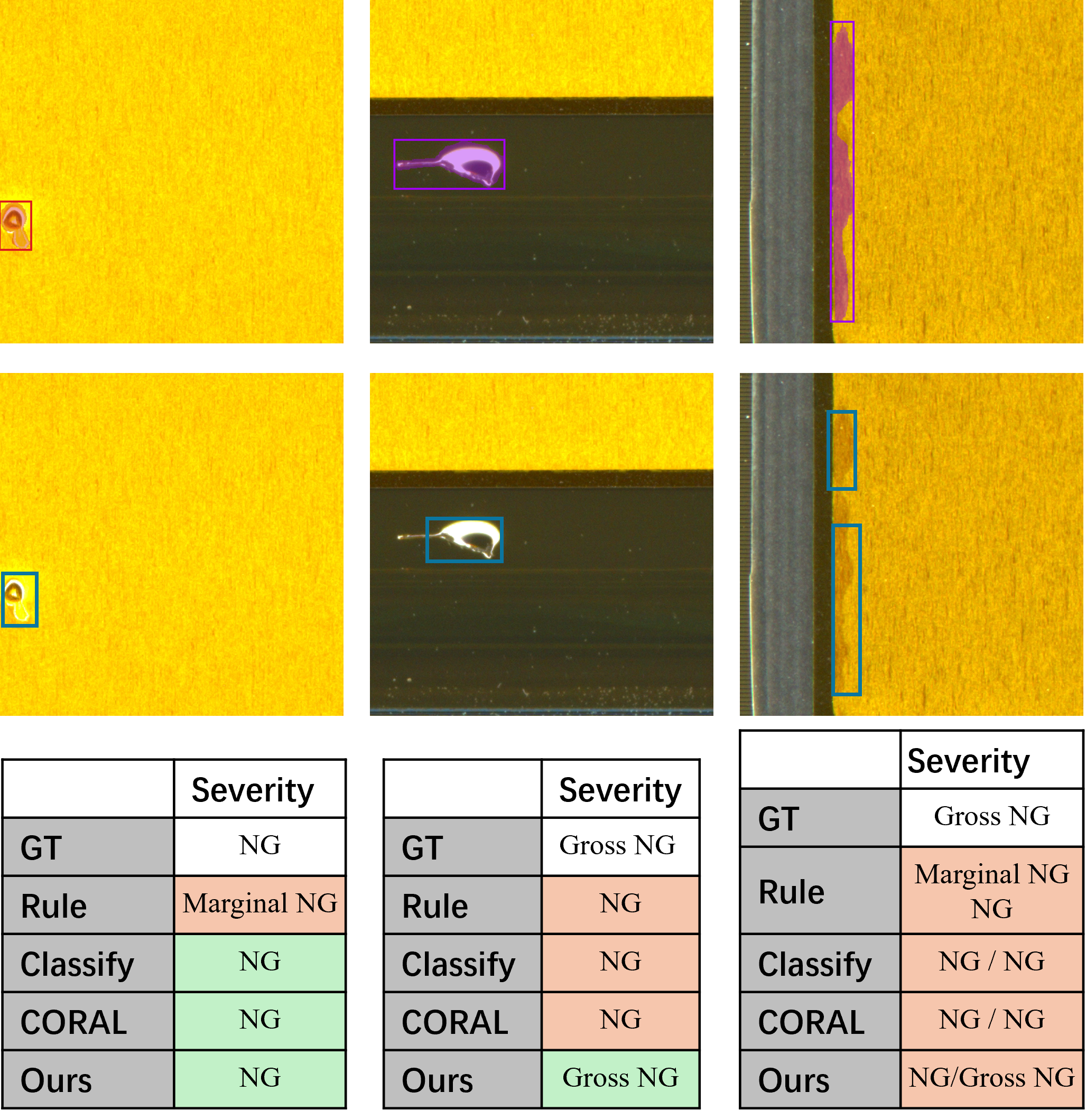}
\caption{Qualitative comparison under predicted ROI inputs. Row 1: original PCB defect images with GT boxes and masks. Row 2: the same images with detector-predicted boxes. Row 3: results of different methods based on the corresponding predicted ROIs.}
\label{fig_4}
\end{figure}

Fig.~\ref{fig_4} shows a qualitative comparison under clean and detector-generated ROI conditions. The first row presents the original PCB images with ground-truth annotations, the second row shows detector-predicted boxes, and the third row compares severity results based on predicted ROIs. In the first example, the predicted box is close to the ground truth, and most learning-based methods produce correct predictions, while the rule-based method fails, indicating its limited robustness. In the second example, larger localization errors introduce ROI contamination; only MAOL correctly predicts \emph{Gross NG}, whereas other methods underestimate it, demonstrating the advantage of our method under noisy conditions. The third example illustrates a severe failure case, where the defect is split into multiple predicted boxes, leading to incorrect predictions for all methods. This highlights a fundamental limitation of the two-stage paradigm, as performance depends on detection quality. Nevertheless, MAOL produces the least severe error, indicating improved robustness compared to other approaches.

\begin{table}[t]
\centering
\caption{Comparison of enhanced ordinal grading models built on top of the CORAL baseline under the predicted-instance setting.}
\label{tab:enhanced_models}
\begin{tabular}{lccc}
\toprule
Strategy & Acc. & Macro Acc. & QWK\\
\midrule
CORAL baseline & 55.8 & 42.1 & 51.1\\
+ Morphology & 72.5 & 69.9 & 76.4\\
+ Class embedding & 75.9 & 72.8 & 80.5\\
+ Adaptive thresholds & 76.5 & 75.8 & 84.2\\
+ Perturbation (ours) & 84.8 & 81.2 & 88.9\\
\bottomrule
\end{tabular}
\end{table}

\subsection{Ablation Study and Threshold Analysis}
\subsubsection{Ablation Study}
To analyze the contribution of each component, we conduct a progressive ablation study starting from the CORAL baseline, as shown in Table~\ref{tab:enhanced_models}. Components are added sequentially, including the morphology branch, class embedding, adaptive thresholds, and perturbation training.

Several observations can be made. First, the morphology branch yields the largest improvement (about +25 QWK), highlighting the importance of explicit structural cues, which are not fully captured by ROI appearance features. Second, the class embedding provides additional but moderate gains, mainly serving as category-aware context for adaptive threshold modeling. Third, adaptive thresholds further improve performance, confirming that severity boundaries are defect-dependent rather than globally shared. Finally, perturbation training brings additional gains by reducing the train--test discrepancy between clean and predicted ROIs, improving robustness in practical deployment.

Overall, the components are complementary: morphology provides the dominant gain, adaptive thresholds further enhance ordinal modeling, and perturbation training improves robustness under noisy ROI conditions.

\subsubsection{Analysis of Adaptive Thresholds}
\begin{figure}[t]
\centering
\includegraphics[width=3.4in]{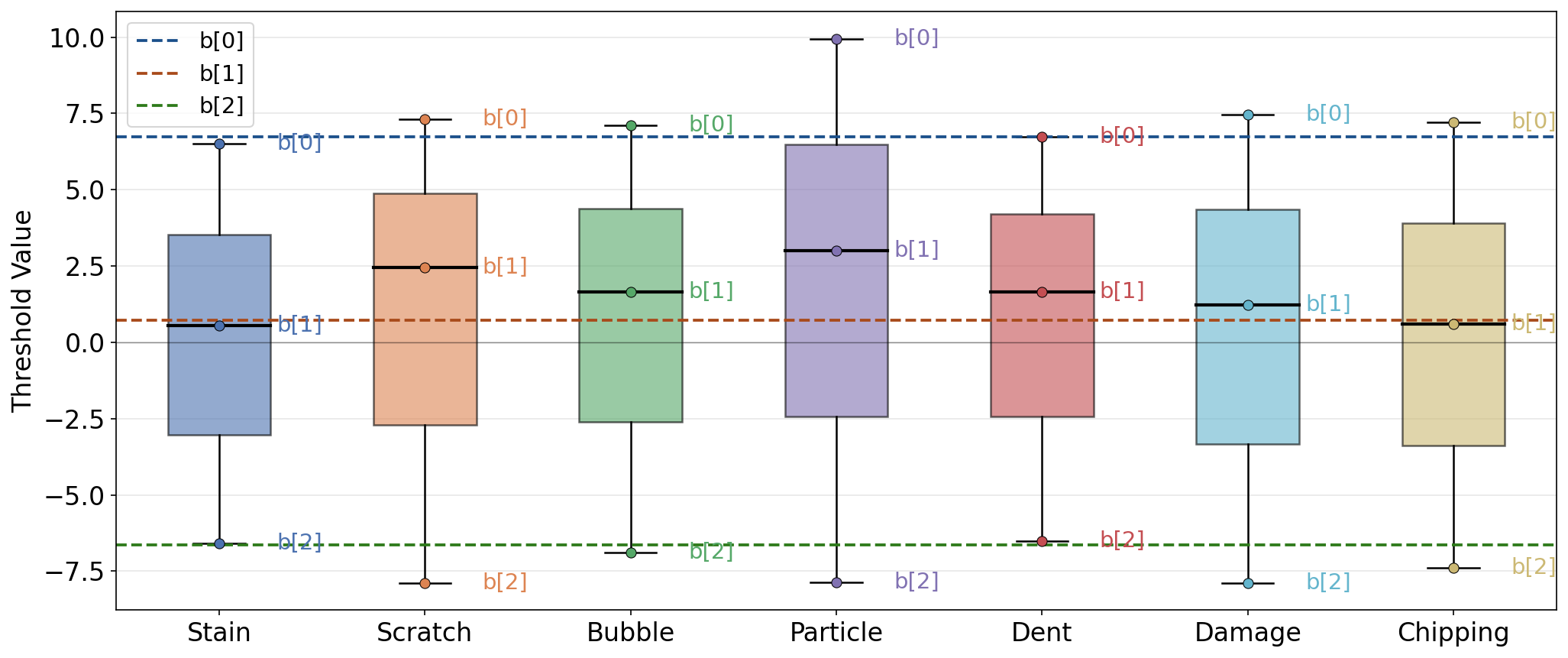}
\caption{Class-specific adaptive ordinal thresholds for seven defect categories. Box plots show the learned values of the three adaptive thresholds for each category, while the horizontal dashed lines indicate the corresponding category-agnostic thresholds learned without adaptive thresholding.}
\label{fig:threshold_boxplot}
\end{figure}

To further analyze the necessity of adaptive thresholds, Fig.~\ref{fig:threshold_boxplot} visualizes the learned class-specific ordinal thresholds for seven defect categories, with global thresholds shown as dashed lines for reference. The results show that thresholds vary significantly across categories, indicating that severity transition patterns are not uniformly shared and cannot be adequately modeled by a single set of global thresholds. Moreover, the relative spacing between thresholds also differs, suggesting that ordinal intervals between severity levels are category-dependent. This implies that the same severity level may correspond to different physical defect extents across categories. Therefore, adaptive thresholds do not simply introduce an additive bias, but effectively reshape category-specific decision boundaries in the latent ordinal space.

\section{Conclusion}
This paper studied fine-grained industrial defect severity grading as an instance-level ordinal prediction problem. Instead of treating severity levels as independent categories, we argued that severity grading should explicitly account for the ordered structure of labels, the morphology-related characteristics of defects, and the train--test mismatch between clean training ROIs and noisy detector-generated instances. To this end, we proposed MAOL, a morphology-aware ordinal learning framework for industrial defect severity grading. The proposed method incorporates three key designs: a morphology branch for explicit structural cues, a class-conditional adaptive ordinal head for category-dependent severity boundaries, and a perturbation-based training strategy to improve robustness under predicted-ROI inputs. Experimental results under both clean-ROI and predicted-instance settings demonstrated that the proposed framework consistently outperforms rule-based, nominal classification, and standard ordinal baselines, with particularly clear advantages in the more practical predicted-instance scenario.

The results validate the importance of combining ordinal modeling, morphology-aware representation, and robustness-oriented training for industrial severity grading. In future work, we will explore end-to-end frameworks that jointly optimize defect localization and ordinal severity grading within a unified model.

\end{document}